\documentclass{article}
\usepackage{PRIMEarxiv}

\usepackage[utf8]{inputenc} 
\usepackage[T1]{fontenc}    
\usepackage{hyperref}       
\usepackage{url}            
\usepackage{booktabs}       
\usepackage{amsfonts}       
\usepackage{nicefrac}       
\usepackage{microtype}      
\usepackage{lipsum}
\usepackage{fancyhdr}       
\usepackage{graphicx}       
\graphicspath{{media/}}     

\usepackage{algorithm}
\usepackage{algorithmic}

\usepackage{graphicx}
\usepackage{microtype}
\usepackage{graphicx}
\usepackage{subfigure}
\usepackage[utf8]{inputenc} 
\usepackage[T1]{fontenc}    

\usepackage{url}            
\usepackage{booktabs}       
\usepackage{amsfonts}       
\usepackage{nicefrac}       
\usepackage{microtype}      
\usepackage{lipsum}
\usepackage{fancyhdr}       
\usepackage{graphicx}       
\graphicspath{{media/}}     

\usepackage{arydshln}  
\usepackage{mathtools}
\usepackage{amsmath,amssymb,amsfonts}

\usepackage{comment}
\usepackage{enumerate}
\usepackage{colortbl}
\usepackage{booktabs}
\usepackage{array}
\usepackage[english]{babel}
\usepackage{amsthm}

\theoremstyle{definition}

\newtheorem{prop}{Proposition}

\newtheorem{definition}{Definition}

\theoremstyle{remark}
\newtheorem{remark}{Remark}

\usepackage{multirow}
\usepackage{adjustbox}

\usepackage{pifont}

\usepackage{xcolor}

\usepackage{newfloat}
\usepackage{listings}
\title{RD-DPP: Rate-Distortion Theory Meets Determinantal Point Process to Diversify Learning Data Samples}

\author{
  Xiwen Chen, Abolfazl Razi \\
  Clemson University  \\
  \texttt{\{xiwenc, arazi\}@clemson.edu} \\
   \And
  Huayu Li \\
  The University of Arizona \\
  \texttt{hl459@arizona.edu}\\
  \And
  Rahul Amin \\
  MIT Lincoln Laboratory \\
  \texttt{Rahul.Amin@ll.mit.edu}\\
}

\usepackage{bibentry}

\begin{document}

\maketitle

\begin{abstract}

In some practical learning tasks, such as traffic video analysis, the number of available training samples is restricted by different factors, such as limited communication bandwidth and computation power. 
Determinantal Point Process (DPP) is a common method for selecting the most diverse samples to enhance learning quality. However, the number of selected samples is restricted to the rank of the kernel matrix implied by the dimensionality of data samples. Secondly, it is not easily customizable to different learning tasks. 
In this paper, we propose a new way of measuring task-oriented diversity based on the Rate-Distortion (RD) theory, appropriate for multi-level classification. To this end, we establish a fundamental relationship between DPP and RD theory. 
We observe that the upper bound of the diversity of data selected by DPP has a universal trend of \textit{phase transition},
which suggests that DPP is beneficial only at the beginning of sample accumulation. This led to the design of a bi-modal method, where RD-DPP is used in the first mode to select initial data samples, then classification inconsistency (as an uncertainty measure) is used to select the subsequent samples in the second mode. This phase transition solves the limitation to the rank of the similarity matrix. 
Applying our method to six different datasets and five benchmark models 
suggests that our method consistently outperforms random selection, DPP-based methods, and alternatives like uncertainty-based and coreset methods under all sampling budgets, while exhibiting high generalizability to different learning tasks
\footnote{The source code is available on \url{https://anonymous.4open.science/r/RD-DPP-83DB}}.

\end{abstract}

\section{Introduction}

A higher data diversity, even in potentially unknown representation space, is known to boost the prediction power of Machine Learning (ML) algorithms. This matter is critical in a class of applications such as Unmanned Aerial Systems (UAS), where data collection capacity is highly constrained by limited power and networking resources. Another example is recommender systems, where more diverse data samples mimic inherent subject-specific variations represented by a complex distribution geometry.

A powerful tool to enhance diversity is the Determinantal Point Process (DPP)~\cite{kulesza2012determinantal,derezinski2019fast,calandriello2020sampling}, which offers a formal approach to model diversity by quantifying dissimilarity among elements within a set, potentially in some latent feature space.
It is widely used by the \textit{machine learning} community in search engines, recommender systems \cite{chen2018fast}, document summarization \cite{perez2021multi}, and more recently in learning-based image processing \cite{launay2021determinantal} and regression models \cite{tremblay2019determinantal,derezinski2021determinantal}.

A related concept is Rate-Distortion (RD) theory commonly used by the \textit{information theory} community to design and evaluate Source Codes (SC) for lossy data compression \cite{cover1999elements}. It characterizes the minimum compression rate for a tolerable distortion level based on the distribution geometry of data samples.

In this research, we reveal the inherent relationship between DPP and RD theory. The relation between RD and DPP comes from the fact that both methods are used to evaluate data diversity but from different perspectives. DPP evaluates the diversity by modeling the dissimilarity among samples in a set, while RD quantifies the minimum representation bits per sample (i.e. the compressibility of the samples) required for a given distribution to satisfy a certain distortion limit. Therefore, they are intrinsically related. This relationship is not received the deserved attention from the research community. This study uses this fundamental relationship to design a new data selection policy.

\begin{figure*}[]
\centering\includegraphics[width=0.82\textwidth]{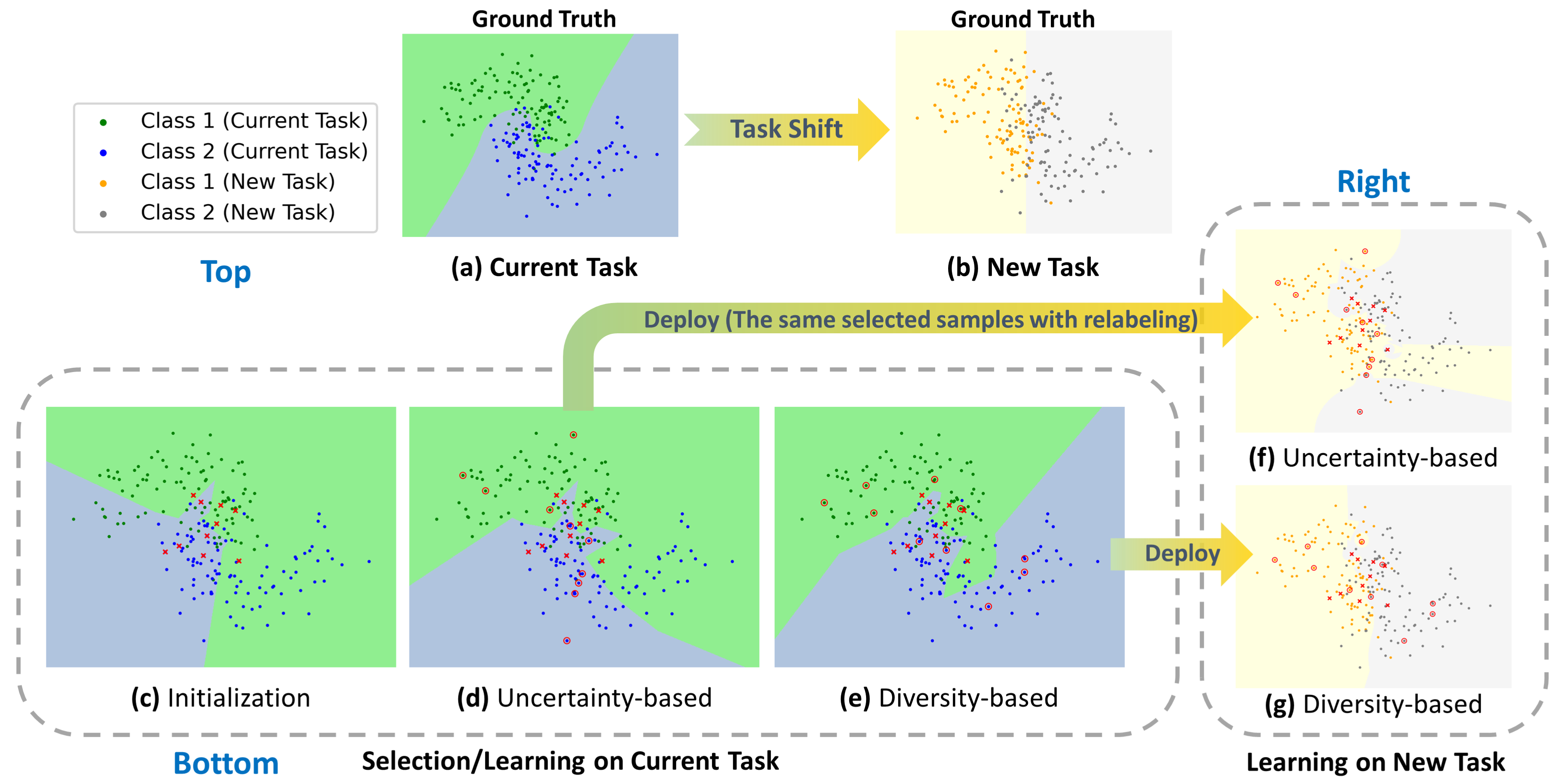}
\caption{The benefit of using diversity-based methods in low-budget conditions; \textbf{Top:} ground truth decision boundary is shown for the \textit{current task} (a) and \textit{new task} (b). \textbf{Bottom:} (c-e): decision boundary learned using a kNN classifier for three scenarios using 10 initial random samples, marked as 'x', in (c); 10 initial samples + 10 uncertainty-based samples (selected based on (c), marked as 'o') in (d), and 10 initial samples + 10 diversity-based samples (selected based on (c), marked as 'o') in (e).  
The diversity-based method (e) is superior for mimicking distribution geometry than the uncertainty-based method. 
\textbf{Right:} compares generalizability of different methods by applying the selected samples for the \textit{current task} to the \textit{new task}. The diversity-based method (e$\rightarrow $g) that captures the overall geometry features is more generalizable than the uncertainty-based method (d$\rightarrow $f), which excessively focuses on specific decision boundaries. }
\label{fig:div_sat}
\end{figure*}

The known drawback of diverse sample selection by conventional DPP is that the number of selected samples cannot be larger than the rank of the kernel matrix (i.e. if $r$ is the rank of the similarity matrix, the probability of selecting subsets with cardinality above $r$ is 0 by DPP \cite{horn2012matrix}). Another consideration is that DPP is not task-oriented and considers merely the inherent diversity of data samples; therefore, data selected based on DPP may not necessarily yield the highest performance for different learning tasks. For instance, selecting diverse samples near decision boundaries (e.g., SVM support vectors) is advantageous for multi-level classification.
To address this concern, one may measure diversity in a latent space, like the ones captured by the later layers of utilized Deep Learning (DL) architectures \cite{yu2020learning,chu2023image}.
In these situations, applying DPP to the representation of data samples obtained by DL architectures can be used to collect samples sequentially, but the issue is that the network weights get updated, and hence the data representation changes by adding new samples.
Therefore, DPP is more appropriate for raw data samples or for one-shot inference, and not for online learning tasks. 
An alternative approach to data selection is using uncertainty-based methods by selecting data samples that are less consistent with the trained model based on metrics like \textit{cross-entropy} and \textit{margin}~\cite{jiang2019minimum,scheffer2001active}.
An issue with this approach is its sensitivity to initial samples, causing poor early-stage performance until sufficient diverse samples are collected to establish reliable decision boundaries.

\begin{figure}[htbp]
\centering\includegraphics[width=0.4\columnwidth]{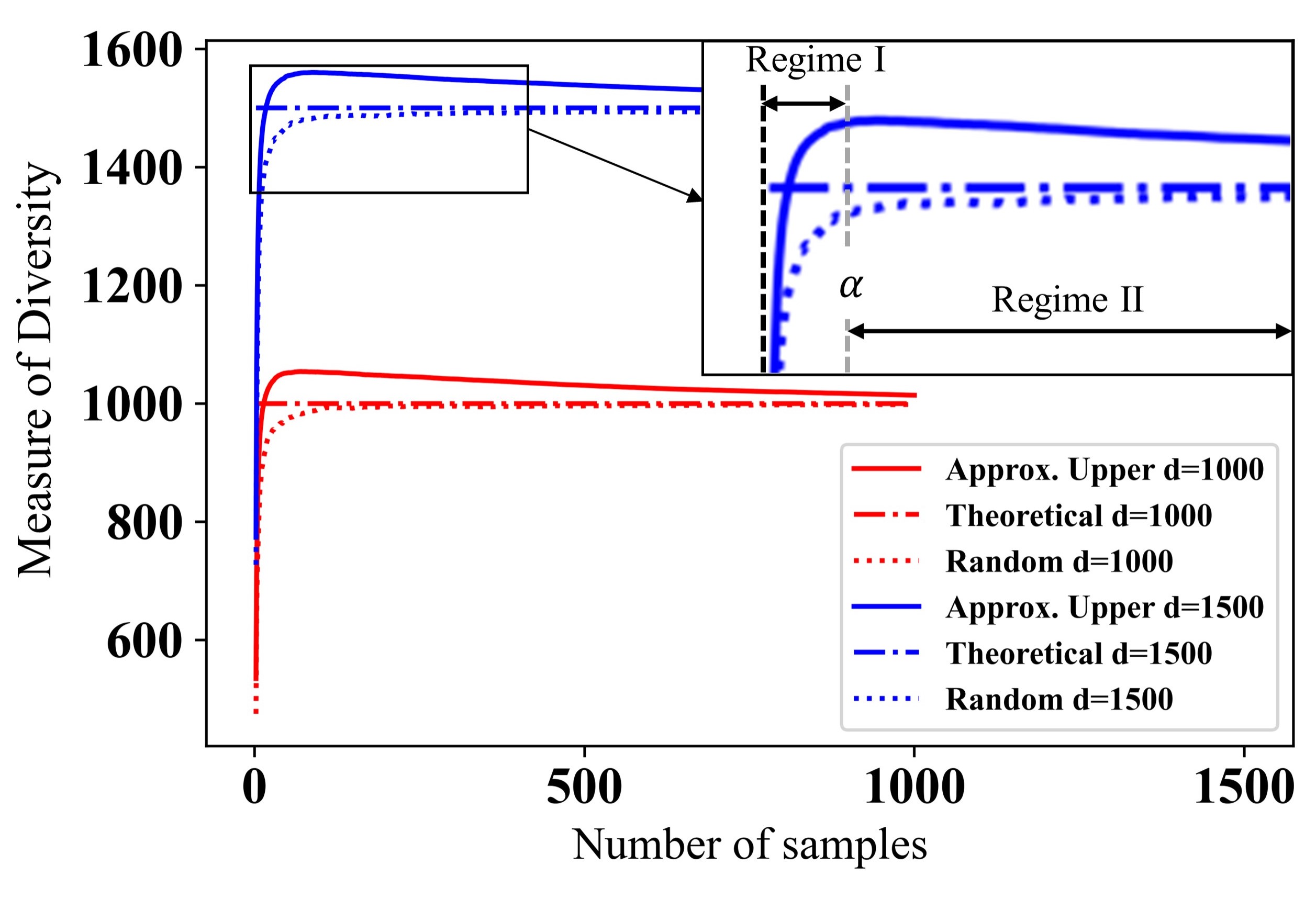}
\caption{Approximate RD-based measure of diversity w.r.t number of selected samples for greedy selection of samples using DPP (solid line) and random selection (dashed line) for $n=2000, d=1000$ and $n=2000, d=1500$, respectively. Both methods converge to the theoretical limits determined by RD theory. Using DPP is advantageous only for the first few samples until $k$ is less than $\alpha< min(d,n)$.}
\label{fig:trend}
\end{figure}

Considering these two perspectives, we develop a new bi-modal algorithm called RD-DPP that facilitates sequential data selection. In the first mode, we use class-conditional RD to measure the task-oriented semantic diversity (e.g., for a classification task) and perform Maximum a Posteriori (MAP) inference for DPP with RD-based quality-diversity kernel. The quality score of the kernel quantifies the added diversity of the sets with new samples with respect to the previously selected samples. In the second mode, we use uncertainty methods to collect the most useful samples.
Indeed, our architecture is consistent with the following observation. 
We observe that the upper bound of diversity reaches its maxima and then gradually converges to its theoretical limit (An example is shown in Fig. \ref{fig:trend}). We conjecture that this \textit{phase transition} is a universal concept.\footnote{Our investigation of different datasets as well as synthetic data with seven different probability distributions in Appendix A supports this observation.}
It means that diversity-based selection is more beneficial at the beginning of data accumulation (the first few samples) to mimic the distribution geometry. After the transition point, the diversity requirement is already fulfilled; hence, using uncertainty-based methods would be equally or even more advantageous.

An illustration of this phenomenon is presented in Fig. \ref{fig:div_sat}, which showcases a non-spherical dataset under budget limitations when the model is built with a few samples selected using the uncertainty-based method (d) and diversity-based method (e). Prioritizing diversity (Fig. \ref{fig:div_sat} (e)) to approximate the population's distribution can lead to a more effective decision boundary compared to emphasizing the reduction of uncertainty of new samples in the close proximity of the decision boundary, especially when the initial model is not reliable. We also found that pursuing diversity is more beneficial for potential future tasks with different ground decision boundaries, as shown in Fig. \ref{fig:div_sat} (f-g). The corresponding experiment is shown in Section \ref{sec:future}.


\paragraph{Contributions:} 
\begin{itemize}
    \item It is the first work to reveal Rate-Distortion theory and DPP are mathematically related for 
Gaussian distributed samples, hence maximizing RD can be used as a diversity tool to measure task-oriented diversity. 
\item We characterize the \textit{phase transition} concept in the diversity-based data accumulation (in Proposition 1), offering insights into data selection strategies. This concept is used to design a bi-modal data selection policy by switching from DPP-based to uncertainty-based selection at the transition epoch.
\end{itemize}
The results in section \ref{sec:exp} show that our method outperforms all alternative methods, including coreset-based methods, two uncertainty-based methods, pure DPP, and random selection by a significant margin.


\begin{prop}
\label{prop:2}
Suppose a set of data points sampled from a certain distribution $\mathbf{z}_1,\mathbf{z}_2,\cdots,\mathbf{z}_n$ with $\mathbf{z}_i\sim P\in \mathbb{R}^{d}$. 
If we use the greedy DPP (e.g., \cite{chen2018fast}) to infer $k$ diverse points to be evaluated by the upper bound of the Rate-Distortion theory, we observe that the $k$ most diverse points are in $[1,\min(d,n)]$, where $d$ is the dimension of data and $n$ is the number of data samples. If $k>\min(d,n)$, the determinant will approach zero). In fact, there is a transition point $\alpha$, so that for $k \in [1,\alpha]$, the upper bound of sample diversity sharply increases to its global maxima, and for $k \in (\alpha,\min(d,n)]$, it slowly declines and converges to the theoretical limit defined by the RD theory. 
\end{prop}
An example of Proposition \ref{prop:2} is shown in Fig. \ref{fig:trend}. We set dimension $d$ to 1000 and 1500 and generate 2000 data points by $\mathbf{z}_i\sim \mathcal{N}(0,\mathbf{I}_d)$, respectively.

\section{Background Information} 
\subsection{Rate-distortion Theory}
An arbitrary real number (e.g., samples of continuous-valued signals) requires an infinite number of bits for lossless representation, which is impractical in most communication and storage systems. 
In practice, we usually settle with lossy compression that allows some representation errors.
Specifically, given an arbitrary source $X$, we can use 
$nR$ bits to encode a sequence of $n$ samples $X^n$ with $f_n(X^n)$ (using a codebook of size $2^{nR}$) and then decode it with 
$\hat{X^n}=g_n(f_n(X^n))$. The reconstruction error for a sample sequence $x^n$ is defined as $d(x^n,\hat{x}^n) := 1/n \sum_{i=1}^n d(x_i,\hat{x}_i)$ for some distance measure $d()$. A commonly used distortion metric is Mean Squared Errors (MSE) $\epsilon^2:=1/n \sum_{i=1}^n (x_i-\hat{x}_i)^2$ and distortion $D$ is defined as $D:=\mathbb{E}[d(X^n, \hat{X}^n)]$ \cite{cover1999elements}.
Rate distortion theory is used to quantify the minimum number of representation bits per sample $R$ for a sequence with infinite length ($n \rightarrow \infty$) and distortion $D$. 
Note that RD is estimated for a finite set of i.i.d. Gaussian distributed samples as follows. 
\begin{definition}\label{def:1}
Let's assume a finite dataset is represented by  $\mathbf{Z}=[\mathbf{z}_1,\mathbf{z}_2,\cdots,\mathbf{z}_n]\in \mathbb{R}^{d\times n}$ (can be in some potentially learnable feature space) presenting $n$ i.i.d. data points, each with $d$ features, sampled from a zero-mean multivariate Gaussian distribution with covariance $\mathbf{\Sigma}$. 
The theoretical coding rate $ R(\mathbf{Z}, \epsilon):= \frac{1}{2} \log \operatorname{det}\left(\frac{d}{ \epsilon^2} \mathbf{\Sigma}\right)$ for a very small tolerable distortion $\epsilon^2$ in squared error (SE) sense, can be approximately estimated 
as \cite{ma2007segmentation}
\begin{align}\label{eq:r0}
    R(\mathbf{Z}, \epsilon) &:= \frac{1}{2} \log \operatorname{det}\left(\mathbf{I}+\frac{d}{n \epsilon^2} \mathbf{Z} \mathbf{Z}^{\top}\right),
\end{align}
where the unit of $ R(\mathbf{Z}, \epsilon)$ is bit/dimension for log base $2$. 
Note that according to the \textit{Hadamard's inequality}~\cite{petersen2008matrix}, we have
\begin{align}\label{eq:r1}
&    \frac{1}{2} \log \operatorname{det}\left(\mathbf{I}+\frac{d}{n \epsilon^2} \mathbf{Z} \mathbf{Z}^{\top}\right) \\ 
&\leq \underbrace{\sum_{i=1}^d \log (\frac{d}{\epsilon^2}\frac{( \mathbf{Z} \mathbf{Z}^{\top})_{ii}}{n}+1)
    \approx {\sum_{i=1}^d \log (\frac{d}{\epsilon^2}\sigma_i^2+1)}}_{\text{upper bound}}, \nonumber
\end{align}
\end{definition}
where $(\mathbf{X})_{ij}$ represents an element of $\mathbf{X}$ in row $i$ and column $j$.

\begin{remark}
We use the term \textit{approximate diversity} instead of \textit{rate} when using metric $R(\mathbf{Z},\epsilon)$ to emphasize that it is an approximate empirical measure computed for a finite set without using the parameters of its distribution.
\end{remark}


Note that for labeled data, each class can be compressed/encoded separately.
\begin{definition}
 The coding rate of the sub-space for each class $R^c_i(\mathbf{Z}, \epsilon \mid C_i)$ is given by,
\begin{align}
       R^c_i(\mathbf{Z}, \epsilon \mid{C}_i) :=\frac{1}{2} \log \operatorname{det}\left(\mathbf{I}+\frac{d}{|C_i| \epsilon^2} \mathbf{Z}_{C_i} \mathbf{Z}^{\top}_{C_i}\right) ,
\end{align}
where $C_i$ is the index set of class $i$,
$c_T$ is the number of classes, $\mathbf{Z}_{C_i}$ is a matrix using columns of $\mathbf{Z}$ indexed by $C_i$ ($\mathbf{Z}[:,C_i]$), and $|C_i|$ is the cardinality of $C_i$. 
\end{definition}

\subsection{Determinantal Point Processing}
\begin{definition}\label{def:2}
DPP is a probability measure on all $2^{|\mathcal{A}|}$ subsets of $\mathcal{A}$, where $|\mathcal{A}|$ denotes the cardinality of the set $\mathcal{A}$. 
According to the definition of DPP \cite{kulesza2012determinantal}, an arbitrary subset ${A}\subseteq \mathcal{A} $ drawn from $\mathcal{A}$ must satisfy,
\begin{align}
\mathcal{P}({A})\propto\operatorname{det}\left(\mathbf{L}_{{A}}\right).
\end{align}%
\end{definition}

Here, $\mathbf{L}$ is a positive semi-definite (PSD) \textit{Gram Matrix} defined as $\mathbf{L}=\mathbf{Z}^{\top} \mathbf{Z}$ to measure pairwise similarity among points, and $\mathbf{L}_{{A}}$ is a submatrix of $\mathbf{L}$ with rows and columns indexed by set $A$.  
We need normalization factor $\operatorname{det}(\mathbf{L}+\mathbf{I})$ when computing exact probabilities, because 
\begin{align}\label{eq:dpp}
     \sum_{{A}\subseteq \mathcal{A}} \operatorname{det}\left(\mathbf{L}_{A}\right) = \operatorname{det}\left(\mathbf{L}+\mathbf{I}\right),
\end{align}
where $A$ denotes a subset drawn from the entire set $\mathcal{A}$, for any ${A}\subseteq \mathcal{A}$.
This identity has been proved in multiple materials, such as \cite{kulesza2012determinantal}. For data selection purposes, we often expect to ensure $k$ samples with the largest diversity. This problem is known as Maximum a Posteriori (MAP) inference for DPP presented as,
\begin{align}\label{eq:map}
        \max_{A\subseteq \mathcal{Z}}& ~\text{det} (\mathbf{L}_A), \quad
        s.t. ~|A| = k, ~k\leq \text{rank}(\mathbf{L}). 
    \end{align}
It is an NP-hard problem, and the common solution is using \textit{greedy search}. A recent fast-known exact greedy approach was proposed in \cite{chen2018fast}, and we denote it as $A^* = DPP_m(\mathbf{L},k)$.

\section{Methodology}
\paragraph{Relation Between Rate-Distortion Theory and DPP:}\label{sec:relations}
Rate-distortion Theory and DPP are inherently related,  as we present here.
Let's set $\alpha :=\frac{d}{n \epsilon^2}>0$ in Eq. (\ref{eq:r0}). The value of the approximate diversity by RD theory (presented in Definition \ref{def:1}) can be described by the sum of the subset probabilities measured by DPP (presented in Eq. (\ref{eq:dpp})), as follows 

\begin{align}
\nonumber
    R(\mathbf{Z}, \epsilon)& \stackrel{(a)}{=} \frac{1}{2} \log \operatorname{det}\left(\mathbf{I}+\alpha \mathbf{Z} \mathbf{Z}^{\top}\right)  \stackrel{(b)}{=} \frac{1}{2} \log \operatorname{det}\left(\mathbf{I}+\alpha \mathbf{Z}^{\top} \mathbf{Z}\right) \\ 
    & \stackrel{(c)}{=} \frac{1}{2} \log \sum_{{X}\subseteq \mathcal{Z}} \operatorname{det}\left(\mathbf{L}_{X}\right),
    \end{align}
where $\mathbf{L}=\alpha \mathbf{Z}^{\top} \mathbf{Z}$ can be viewed as the L-ensemble kernel matrix of DPP, and $\mathcal{Z}=\{1,2,\cdots,n\}$ denotes the index set of $\mathbf{Z}$.  
This relation states that the $ R(\mathbf{Z}, \epsilon)$ can be described as the sum of point process measurements $\operatorname{det}\left(\mathbf{L}_{X}\right)$, which reveals a diverse set of samples should have high diversity for all of its possible subsets. For the proof of Eq. (7), please refer to Appendix B.

\paragraph{DPP Approaches to Solve RD Problem:}\label{sec:RD-DPP}
Based on the inherent relationship between the DPP and RP, we develop an RD-based quality function to measure individual rate gain as follows. 
Given a previously selected data set $\mathbf{Z}\in \mathbb{R}^{d\times n}$, how to search a new sample set $\mathbf{D} = \{\mathbf{z}_{d_1}, \mathbf{z}_{d_2}, \cdots \mathbf{z}_{d_k}\} $ with indices $\mathcal{D}=\{d_1,d_2,\cdots,d_k\}\subseteq \mathcal{B}$ such that the resulting diversity of $\mathbf{Z}^{\mathcal{D}+} := [\mathbf{Z},\mathbf{z}_{d_1}, \mathbf{z}_{d_2}, \cdots \mathbf{z}_{d_k}]\in \mathbb{R}^{d\times (n+k)}$ (i.e. the diversity measured by Definition \ref{def:1}) is maximized, where $\mathcal{A}$, $\mathcal{Z}\subset \mathcal{A}$, and $\mathcal{B} = \mathcal{A} \setminus \mathcal{Z}$, respectively, denote the index set of entire data, previously selected samples, and candidate samples. Based on our conclusion in Eq. (7), we can compute the diversity after selecting the set $\mathcal{D}$ as 
\begin{align}\label{eq:8}
     R(\mathbf{Z}^{\mathcal{D}+}, \epsilon)&=\frac{1}{2} \log \operatorname{det}\left(\mathbf{I}+\frac{d}{(n+k) \epsilon^2} {\mathbf{Z}^{\mathcal{D}+}}^\top \mathbf{Z}^{\mathcal{D}+}\right) \\ \nonumber
     &=\frac{1}{2} \log\sum_{{X}\subseteq \mathcal{Z}^{\mathcal{D}+}}\operatorname{det}\left(\widetilde{\mathbf{L}}_{X}\right),
\end{align}
where $\widetilde{\mathbf{L}}=\frac{d}{(n+k) \epsilon^2}\mathbf{Z}^{\top}  \mathbf{Z}$. $\widetilde{\mathbf{L}}_X$ is the submatrix of $\widetilde{\mathbf{L}}$ indexed by $X$, and $\mathcal{Z}^{\mathcal{D}+}= \mathcal{Z}\cup \mathcal{D}$ is the index set of $\mathbf{Z}^{\mathcal{D}+} $.
Our goal here can be stated as
\begin{align}\label{eq:9}
    \arg\max_{\mathcal{D}}& ~ R(\mathbf{Z}^{\mathcal{D}+}, \epsilon), \quad s.t. ~ |\mathcal{D}| =k.
      \setlength{\abovedisplayskip}{5pt}
\end{align}
Since in Eq. (\ref{eq:8}), the logarithm base is 2, we can obtain $ 2^{2R(\mathbf{Z}^{\mathcal{D}+}, \epsilon)} =\sum_{{X}\subseteq \mathcal{Z}^{\mathcal{D}+}}\operatorname{det}\left(\widetilde{\mathbf{L}}_{X}\right)$, which is the sum of DPP-based measure across the all subsets $X\subseteq\mathcal{Z}^{\mathcal{D}+}$. 
We can obtain the similar term $2^{2R(\mathbf{Z}^{b^i+}, \epsilon)}$ for each candidate $b^i\in\mathcal{B}$ as $\sum_X\operatorname{det}\left(\mathbf{L}_{X}\right)$. Here, $X\subseteq\mathcal{Z}^{b^{i+}}:=\mathcal{Z}\cup \{b_i\}$, and we have $({n+k})\widetilde{\mathbf{L}}_{X} =({n+1}){\mathbf{L}}_{X}$. Noting that $R(\mathbf{Z}^{b^i+}, \epsilon)$ is the individual diversity gain for $b^i$ that has the memory of the selected data set $\mathbf{Z}$ but has no knowledge about other candidates, which cannot facilitate diversity among candidates.
By applying the set operation, we can translate the problem to one that considers both the individual rate gain of each candidate and the diversity among candidates. To this end, we develop the following approach to solve the optimization problem in Eq. (\ref{eq:9}). The diversity vector is the feature of each sample $\mathbf{z}_i,i\in \mathcal{B}$, and the quality score $\Phi()$ evaluates the individual rate gain from the perspective of RD theory. 
We can use MAP inference for DPP with a quality-diversity kernel $\mathbf{K}$ to solve the problem  (i.e. $\arg\max_{\mathcal{D}}\mathbf{K}_\mathcal{D}$, where $\mathbf{K}_\mathcal{D}$ is $\mathbf{K}$'s rows and columns indexed by $\mathcal{D}$.)  as follows, 
\begin{align}\label{eq:rdkernel}
    \mathbf{K}_{i,j}=\Phi(\mathbf{Z}^{b^{i+}}, \epsilon)\Phi(\mathbf{Z}^{b^{j+}}, \epsilon)(\mathbf{L}_\mathcal{B})_{i,j}
\end{align}
where $\mathbf{L}_{\mathcal{B}}=\mathbf{Z}_{\mathcal{B}}^{\top}\mathbf{Z}_{\mathcal{B}}$ 
is the gram matrix across all $m$ candidate samples with index $\mathcal{B}=\{b^{1},b^{2},\cdots,b^{m}\}= \mathcal{A}  \setminus \mathcal{Z}$, $\mathbf{Z}^{b^{i+}} := [\mathbf{Z},\mathbf{z}_{b^{i}}]$. $\Phi(\mathbf{Z}^{b^{i+}})$ is the RD-based quality function to quantify the individual gain obtained by adding this sample to the known set. 
\begin{figure}[htbp]
\centering\includegraphics[width=0.5\columnwidth]{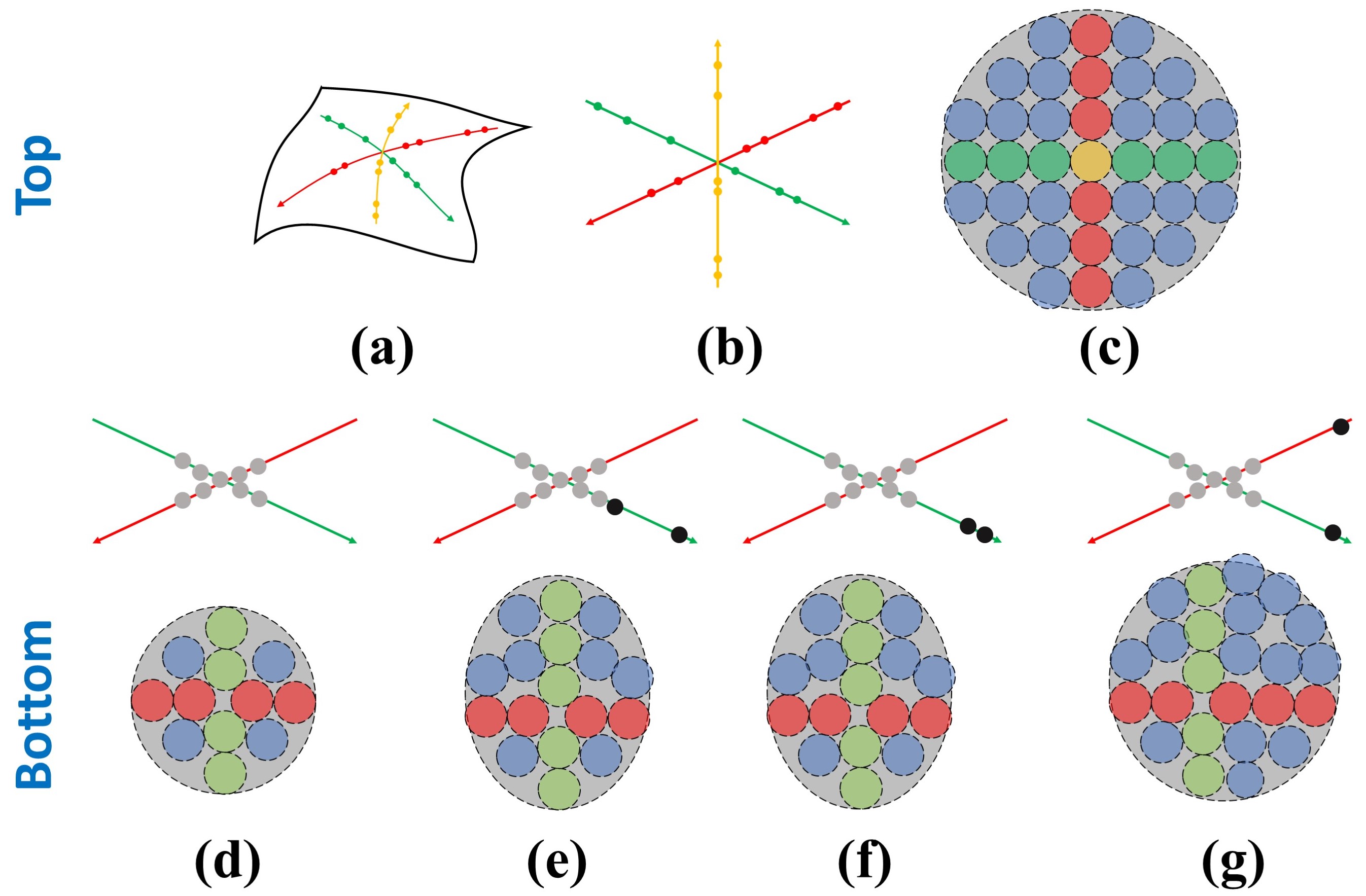}
\caption{Visualizing the concept of task-oriented diversity. \text{Top:} Green, red, and yellow denote three different classes: (a) high-dimensional data, (b) corresponding low-dimensional manifold, and (c) \textit{sphere packing}, where each sphere denotes a bit with $\epsilon^2$ distortion tolerance. Here, yellow spheres are orthogonal on the plane. 
\text{Bottom:} Different scenarios for adding two new samples (black points) to the previously selected set of nine samples (gray points). 
(d) previously selected samples; (e-g) adding two samples based on the pure diversity of new samples (e), the individual marginal gain of the RD-based diversity (f). and the highest semantic diversity (g).}
\label{fig:low_dim}
\end{figure}
\paragraph{Task-oriented RD-based Kernel:} To enhance the quality of training for a specific learning task, such as the classification task, we develop a \textit{semantic diversity} kernel instead of the original class-independent DPP diversity. 
First, we adopt the assumption by \cite{yu2020learning}: i) The distribution of any high-dimensional data (Fig. \ref{fig:low_dim}(a)) is typically supported on a low-dimensional manifold (Fig. \ref{fig:low_dim}(b)). ii) A good data representation for a classification task should have \textit{within-class diversity} and \textit{between-class discrimination}. This is visualized by sphere packing in Fig. \ref{fig:low_dim}(c), where we expect the number of blue spheres to be large (maximize the between-class discrimination), and so is the number of the color spheres (except blue) to maximize the within-class diversity. Therefore, we can define semantic diversity for a set of samples $\mathbf{X}$ by class-conditional RD as 
\begin{align}\label{eq:div}
    sdiv(\mathbf{X}) :=  R(\mathbf{X}, \epsilon)- \sum_{i=1}^{c_T}\frac{|C_i|}{n} {R^c_i(\mathbf{X}, \epsilon \mid {C}_i)}\geq 0.
\end{align}
where, again, $C_i$ and $c_T$ denote the index set of data in class $i$ and total number of classes, respectively. Likewise, suppose a previously selected data set $\mathbf{Z}$. Like the Eqs. (8)-(10), to maximize the semantic diversity ($sdiv(\mathbf{Z}^{\mathcal{D}+}, \epsilon)$) by selecting an additional set $\mathcal{D}$ ($|\mathcal{D}|=k$), we can apply the RD-DPP relations and develop the RD-based quality function to evaluate the semantic diversity gain caused by selecting individual candidate $\mathbf{x}_{i}, i\in\mathcal{B}$ as,
\begin{equation}\label{eq:qualityfunc}
    \Phi(\mathbf{X}_{i+}, \epsilon) = sdiv(\mathbf{X}_{i+}),
\end{equation}
where $\mathbf{X}_{i+}= [\mathbf{Z},\mathbf{x}_{i}]\in \mathbb{R}^{d\times (n+1)}$.
Then, similar to Eq. (\ref{eq:rdkernel}), the task-oriented DPP kernel $\mathbf{K}$ can be constructed based on Eqs. (\ref{eq:rdkernel}) and (\ref{eq:qualityfunc}), as
\begin{equation}\label{eq:sim}
\setlength{\abovedisplayskip}{3pt}
   \mathbf{K}_{i,j} = \Phi(\mathbf{X}_{i+}, \epsilon) \Phi(\mathbf{X}_{j+}, \epsilon)  \langle \mathbf{x}_{i}, \mathbf{x}_{j}\rangle, 
    \setlength{\belowdisplayskip}{3pt}
\end{equation}
where $\langle \cdot \; , \; \cdot \rangle $ denotes the inner product operation.
Fig. \ref{fig:low_dim}(d)-(g) demonstrates different strategies to add two points to a known set of points for a labeled dataset, which indicates that we should take into account both individual diversity gain and the distance between the two candidates.
A fast DPP MAP inference proposed in \cite{chen2018fast} can be used to search the $k$ optimized candidates as $\arg\max_\mathcal{D}\mathbf{K}_\mathcal{D}$.

\paragraph{Bi-Modal Scheduling:}
To accommodate \textit{phase transition}, we can set an empirical criterion to switch mode from RD-DPP diversity to uncertainty-based selection. 
To this end, we calculate the semantic diversity $sdiv(\mathbf{Z}^{t})$ at the end of each round (for $t=k,2k,3k,\cdots$), and switch when we first observe $sdiv(\mathbf{Z}^{t})-sdiv(\mathbf{Z}^{t-k})<\phi_0$ meaning that the diversity improvement is less than a pre-defined threshold, $\phi_0$.
A summary of our bi-modal algorithm is presented in Algorithm 1.

\begin{algorithm}[t]\label{alg:1}
\caption{Bi-modal RD-DPP for Sample Selection} 
\textbf{Input}: Entire data with indices $\mathcal{A}$, Initial data $\mathbf{Z}_0$ with index set $\mathcal{Z}_0$, affordable transmission budget $n_T$ samples, 
and the number of samples 
$k$ selected in each round. \\
\textbf{Output}: The index set of selection $SelSet$.
\begin{algorithmic}[1]
\STATE \noindent \textbf{Initialize:} $SelSet\leftarrow \mathcal{Z}_0$, $ \mathcal{B}\leftarrow \mathcal{A}\setminus\mathcal{Z}_0$, $\mathbf{Z}\leftarrow \mathbf{Z}_0$, and $t\leftarrow0$, transitionFlag$\leftarrow False$.
\WHILE{$t\leq n_T$}
\STATE $t \leftarrow t+k$ \texttt{\#Mode one. DPP-based.}
 \IF{$sdiv(\mathbf{Z}^{t})-sdiv(\mathbf{Z}^{t-k})>\phi_0$ \textbf{and Not} transitionFlag} 
 \STATE Calculate the DPP kernel $\mathbf{K}$ for $\mathcal{B}$ by Eq. (\ref{eq:sim}).
\STATE $SelSet\_round \leftarrow DPP_m(\mathbf{K},k) $.
 \ELSE 
\STATE  transitionFlag$\leftarrow True$ \texttt{\#To Mode two}.
\STATE $SelSet\_round\leftarrow Uncertainty(\mathbf{x}_{i}, i\in\mathcal{B}, k)$
\ENDIF
\STATE $\mathcal{B} \leftarrow \mathcal{B}\slash SelSet\_round$  
\STATE $SelSet \leftarrow SelSet\cup SelSet\_round$
 \FOR {$i$ \textbf{in} $SelSet\_round$}
\STATE $\mathbf{Z}\leftarrow[\mathbf{Z},\mathbf{x}_{i}] $. \texttt{\#Add one column to $\mathbf{Z}$.}
\ENDFOR
\ENDWHILE

\end{algorithmic}
\end{algorithm}

\section{Experiment}\label{sec:exp}
We evaluate the proposed and alternative method using six datasets, including MNIST \cite{deng2012mnist}, FMNIST \cite{xiao2017fashion}, CIFAR10 \cite{krizhevsky2009learning}, Yeast \cite{misc_yeast_110}, Cardiotocography \cite{misc_cardiotocography_193}, and Statlog (Landsat Satellite) \cite{misc_statlog}. We compare our method against multiple alternative selection policies, including i) \textit{Uncertainty Decision} (\textit{Uncertainty Dec.}): the selection uses cross-entropy of the predicted labels of new samples are used as the uncertainty metric (higher is more uncertain), ii) \textit{Min Margin Decision} (\textit{Min Margin Dec.} ): selection based on classification margin defined as the difference between the softmax probability of the highest predicted class and the second highest predicted class (less is more uncertain) \cite{jiang2019minimum,scheffer2001active}. For the sake of completeness, we also compare it against two diversity-based methods that use coresets as a representation of the entire set. These methods include iii) DPP Coreset \cite{tremblay2019determinantal}, which is purely based on DPP, iv) K-Center Coreset \cite{sener2017active}, which captures the geometric structure of the original dataset by solving the K-Center clustering problem. Lastly, we examine v) Random selection for our analysis. 
 It is worth mentioning that in practical systems, the data is transmitted in terms of packets that may contain more than one sample. To accommodate this consideration, we regulate the selection of packets (instead of samples) in our experiments. All the above derivations are valid with only one change that the sample's feature vectors $\mathbf{x}$ are replaced by the class-wise mean of feature vectors of all samples within the packet. Details of all experiment setups are provided in Appendix C.

\begin{figure}[]
\centering\includegraphics[page=1,width=0.8\textwidth]{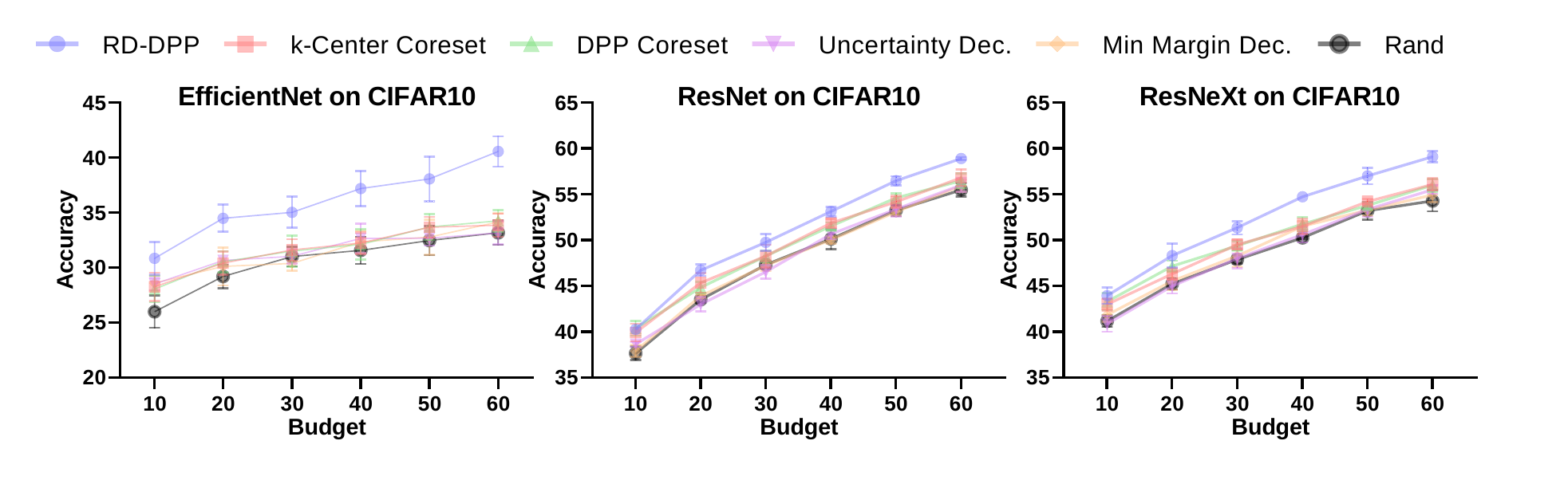}
\caption{Comparing the proposed RD-DPP against K-Center Coreset, DPP Coreset, Uncertainty-based, Min-Margin, and random selection methods applied to the CIFAR10 dataset using three different network architectures: EfficientNet,  ResNet, and  ResNeXt.} 
\label{fig:result_all}
\end{figure}

\paragraph{Experiment on Latent Space:} We first apply a simple CNN network (3 convolutional layers with a fully-connected layer) with random initialization to the MNIST and FMNIST datasets and use the feature from the layer before the classifier as the lower-dimensional representation of data samples (each image is mapped to a $288\times 1$ vector). We assume there exist a total of 100 packets for MNIST and FMNIST, and each packet contains 64 samples. In each experiment, we use 5 randomly selected packets for initialization and then perform different strategies to select packets. The composition of some exemplary packets is shown in Appendix C.1.
\begin{table}[]
\centering
\caption{The classification accuracy (\%) of the trained DL architecture using different selection methods on MNIST and FMNIST datasets. }
\label{tab:result_MF}
\resizebox{0.6\textwidth}{!}{%
\begin{tabular}{lcccccc}\toprule
                          & \multicolumn{3}{c}{\textbf{MNIST}}               & \multicolumn{3}{c}{\textbf{FMNIST}}              \\ \midrule
\textbf{Budget}           & \textbf{10}    & \textbf{30}    & \textbf{50}    & \textbf{10}    & \textbf{30}    & \textbf{50}    \\ \midrule
\textbf{RD-DPP}           & \textbf{49.67} & \textbf{83.21} & \textbf{91.26} & \textbf{44.36} & \textbf{54.75} & \textbf{59.35} \\
\textbf{Uncertainty Dec.} & 22.42          & 70.23          & 90.23          & 27.32          & 51.64          & 57.43         \\
\textbf{Min Margin Dec.}  & 43.14          & 74.14          & 82.4           & 30.39          & 45.14          & 51.12          \\
\textbf{Rand}             & 44.12          & 78.58          & 87.91          & 32.1           & 52.46          & 56.30     \\ \bottomrule     
\end{tabular}%

}
\end{table}
The results (average of 10 runs) are shown in Table \ref{tab:result_MF} and the first row of Fig. \ref{fig:result_all}. Our proposed approach outperforms all other selection methods at any budget. It obtains a $3\%-5\%$ accuracy gain compared to the random selection on MNIST at all budgets from 10 to 50.

For further investigation, we conduct a similar test using the CIFAR10 dataset. Likewise, we construct 100 packets, and each packet has 200 samples. We evaluate our approach with three different state-of-the-art architectures\footnote{The models can be found in \url{https://github.com/kuangliu/pytorch-cifar}.}: EfficientNet-B0 \cite{tan2019efficientnet}, ResNet-18 \cite{he2016deep}, and ResNeXt29 (2x64d) \cite{xie2017aggregated}, respectively, representing each image as a vector of 320, 512, and 1024 elements. 
The results are shown in Table \ref{tab:3net} and the second row of Fig. \ref{fig:result_all}, which exhibit a 
significant gain for our method over the pure uncertainty-based method and random selection. For example, ResNet and ResNeXt, when using our selection strategy obtain a 2\%-3\% gain over 
the uncertainty-based decision, min-margin decision, and random selection. EfficientNet obtains even a higher gain of 5\% over the other methods at transmission budgets up to 50. We also observe that the coreset method and ours outperform the uncertainty-based method at the beginning. However, when the transmission budget increases, 
our method obtains a 2\%-4\% gain over the two coreset methods.





\begin{table}[h]
\centering
\caption{Performance of the classifier (\%) with different network architectures for different selection methods applied to CIFAR10 dataset.}
\label{tab:3net}
\resizebox{0.8\textwidth}{!}{%
\begin{tabular}{lccc|ccc|ccc}\toprule
\textbf{Architecture}     & \multicolumn{3}{c}{\textbf{EfficientNet}}        & \multicolumn{3}{c}{\textbf{ResNet18}}            & \multicolumn{3}{c}{\textbf{ResNeXt}}             \\ \midrule
\textbf{Budget}           & \textbf{10}    & \textbf{30}    & \textbf{50}    & \textbf{10}    & \textbf{30}    & \textbf{50}    & \textbf{10}    & \textbf{30}    & \textbf{50}    \\  \midrule
\textbf{RD-DPP}           & \textbf{30.85} & \textbf{35.03} & \textbf{38.07} & 40.26         & \textbf{49.75} & \textbf{56.49} & \textbf{43.94} & \textbf{51.36} & \textbf{57.02} \\
\textbf{k-Center Coreset} & 28.26          & 31.62          & 33.68          & 39.92          &48.26          & 54.19          & 42.98          & 49.52          & 54.2           \\
\textbf{DPP Coreset}      & 28.05          & 31.52          & 33.69          & \textbf{40.36} & 48.25          & 54.6           & 43.29          & 49.4           & 53.79          \\
\textbf{Uncertainty Dec.} & 28.5           & 31.03          & 32.67          & 38.63          & 46.55          & 53.38          & 40.86          & 48             & 53.36          \\
\textbf{Min Margin Dec.}  & 28.6           & 30.38          & 32.79          & 37.89          & 47.24          & 53.08          & 41.83          & 48.3           & 53.28          \\
\textbf{Rand}             & 25.98          & 31.01          & 32.47          & 37.64          & 47.32          & 53.27          & 41.16          & 47.88          & 53.2          \\ \bottomrule
\end{tabular}%

}
\end{table}


\begin{table}[h]
\centering
\caption{Performance (AUCROC) on three UCI datasets.}
\label{tab:result_small}
\resizebox{0.78\textwidth}{!}{%
\begin{tabular}{lccc|ccc|ccc} \toprule
\textbf{Dataset}            & \multicolumn{3}{c}{\textbf{Yeast}}               & \multicolumn{3}{c}{\textbf{Cardio.}}             & \multicolumn{3}{c}{\textbf{Statlog.}}            \\ \midrule
\textbf{Budget}             & \textbf{3}     & \textbf{9}     & \textbf{15}    & \textbf{3}     & \textbf{9}     & \textbf{15}    & \textbf{3}     & \textbf{9}     & \textbf{15}    \\ \midrule
\textbf{RD-DPP} & \textbf{75.16} & \textbf{81.53} & \textbf{84.33} & \textbf{74.08} & \textbf{85.07} & \textbf{91.47} & \textbf{88.92} & \textbf{94.22} & 95.16          \\
\textbf{Uncertainty Dec.}   & 73.35          & 79.81          & 83.11          & 67.41          & 84.59          & 90.85          & 82.26          & 93.68          & \textbf{95.26} \\
\textbf{Min Margin Dec.}    & 71.29          & 75.97          & 81.36          & 67.57          & 75.9           & 83.42          & 84.13          & 86.48          & 93.9           \\
\textbf{Rand}               & 68.15          & 77.62          & 79.78          & 72.51          & 81.9           & 86.61          & 84.42         & 89.89          & 94.64          \\ \bottomrule  
\end{tabular}%
}
\end{table}

\paragraph{Experiment on Raw Samples:} 
In this respect, we evaluate our method using three UCI small datasets: Yeast \cite{misc_yeast_110}, Cardiotocography \cite{misc_cardiotocography_193}, and Statlog (Landsat Satellite) \cite{misc_statlog}.  Here, we set the total number of packets to 60 and each packet has 5 samples. The selection is initialized with 3 packets, and in each round, we select $k=3$ packets by different approaches. Since their samples are limited, here we only use Logistic Regression as the learning model. To encounter the unbalanced data, we use \textit{Area Under the Receiver Operating Characteristic Curve} (AUCROC) to assess their performances and present the results (average of 10 runs) in Table \ref{tab:result_small}. Again, our proposed selection method outperforms the random selection on Yeast, Cardiotocography, and Statlog datasets with 3\%-7\%, 1\%-4\%, and 0.5\%-4\% gain 
at transmission budgets 3, 9, and 15, respectively.

\paragraph{Ablation Analysis:}
We conducted an ablation study to demonstrate the effectiveness of our algorithm. Two baselines were considered: i) RD-DPP (only diversity), which focuses solely on diversity without transitioning to uncertainty-based methods after diversity reaches the saturation point, 
ii) \textit{Marginal Rate Gain}, which selects the $k$ top candidate samples merely based on their individual semantic gains (i.e. $k$ largest $\Phi(\mathbf{X}_{i+}, \epsilon)$ defined by Eq. (\ref{eq:qualityfunc})) ignoring the within-diversity of the candidate samples. 
The result in Table \ref{tab:ablation} shows that before the transition point, our bi-modal RD-DPP method outperforms the \textit{Marginal Rate Gain} with 8\%-10\% and 5\%-16\% accuracy improvement on MNIST and FMNIST, respectively. Our method is equivalent to RD-DPP (only diversity) before the transition point as expected. 
After the phase transition point (i.e. the point between 20-30 and 40-50 for MNIST and FMNIST, respectively), RD-DPP (bi-modal) consistently achieves around 3\% accuracy gain over the other two baseline methods.

\begin{table}[]
\centering
\caption{The ablation analysis on MNIST and FMNIST, respectively. Here,~ \ding{55}~ denotes there is no \textit{phase transition}, while~\ding{51}~denotes that \textit{phase transition} has occurred in the RD-DPP (Bi-modal). }
\label{tab:ablation}
\resizebox{0.8\textwidth}{!}{%
\begin{tabular}{clcccccc} \toprule
\multicolumn{1}{l}{\textbf{Dataset}} & \textbf{Budget}                  & \textbf{10}    & \textbf{20}    & \textbf{30}    & \textbf{40}    & \textbf{50}    & \textbf{60}    \\ \midrule
\multirow{4}{*}{\textbf{MNIST}}      & \textbf{Phase Trans?}            & \ding{55}            & \ding{55}            & \ding{51}            & \ding{51}            & \ding{51}            & \ding{51}            \\
                                     & \textbf{RD-DPP (Bi-modal)}      & \textbf{49.67} & \textbf{72.86} & \textbf{83.21} & \textbf{89.25} & \textbf{91.26} & \textbf{92.36} \\ \cline{2-8}
                                     & \textbf{RD-DPP (Only Diversity)} & -            & -            & 80.92          & 84.72          & 87.29          & 90.11          \\
                                     & \textbf{Marginal Rate Gain}      & 41.99          & 63.13          & 72.3           & 83.03          & 84.71          & 86.74          \\ \midrule
\multirow{4}{*}{\textbf{FMNIST}}     & \textbf{Phase Trans?}            & \ding{55}            & \ding{55}            & \ding{55}            & \ding{55}            & \ding{51}            & \ding{51}            \\
                                     & \textbf{RD-DPP (Bi-modal)}       & \textbf{44.36} & \textbf{50.75} & \textbf{54.75} & \textbf{55.87} & \textbf{59.35} & \textbf{63.45} \\ \cline{2-8}
                                     & \textbf{RD-DPP (Only Diversity)} & -             & -            & -             & -             & 56.36          & 57.58          \\
                                     & \textbf{Marginal Rate Gain}      & 28.12          & 42.94          & 48.64          & 54.81          & 57.86          & 60.56      \\ \bottomrule   
\end{tabular}%
}
\end{table}


\paragraph{Complexity Analysis:}
In practical systems, especially for image-based learning tasks, the bottleneck is typically the limited transmission budget, and generally, the computation capability of the servers with GPUs is not a concern. Indeed, the DL model is only needed to be trained in the fusion center (the receiver) and the model parameters can be sent back to the source nodes to generate low-dimensional data representations. Still, we calculate the complexity of our model for more clarity. 
In real systems, especially for image-based learning, limited transmission budget is key, while server computation is usually sufficient (including GPUs). 
Our main overhead is to compute the semantic quality score (Eq. (\ref{eq:qualityfunc})). For each candidate $i$, the complexity of the term $ R(\mathbf{X}_{i+}, \epsilon)= \log \operatorname{det}\left(\mathbf{I}+\alpha \mathbf{X}_{i+} \mathbf{X}_{i+}^\top\right)$ is only $\mathcal{O}(\min(t,d )^3)$ (i.e. the complexity of the SVD decomposition of $\mathbf{X}_{i+}$). Therefore, we need operations in the order of $\mathcal{O}(m\min(t,d )^3)$ to compute the semantic quality score of all candidates. Then, constructing the kernel presented in Eq. (\ref{eq:sim}) requires a $\mathcal{O}\big((dc_T)m^2\big) = \mathcal{O}\big(dm^2\big)$ complexity for a small number of cluster/class labels $c_T$. The remaining complexity is the same as the greedy search method \cite{chen2018fast}, which requires a $\mathcal{O}(m^3)$ complexity for initialization and a $\mathcal{O}(k^2m)$ complexity to return $k$ select samples. Thus, the overall complexity in each round is $\mathcal{O}(m\min(t,d )^3+d^2m+m^3+k^2m)\approx \mathcal{O}(m\min(t,d )^3)$. In our work, we use bootstrapping to accelerate the approximation.

\section{Discussion: Learning for Future Tasks}\label{sec:future}
In this section, we highlight the broader advantages of diversity-based selection methods beyond their immediate benefits for current tasks by effectively preserving representative information. 
 
In contrast, uncertainty-based approaches, which are primarily designed to enhance the current model, lack this capacity. Here, we consider two scenarios:

\paragraph{Robustness for Label-shift Generalization:} In this scenario, the same set of selected data samples are used for different tasks. We perform our experiment on the Large-scale CelebFaces Attributes (CelebA) Dataset \cite{liu2015faceattributes}, where each attribute can be used as the target label to perform a binary classification task. 
We select the samples for the \texttt{Smiling} classification task, then train two new classifiers for \texttt{Blond\_Hair} and \texttt{High\_Cheekbones} target labels using the same selected samples. 
The results after running 20 times are reported in Table \ref{tab:multi-task1} for 20, 40, and 60 selected packets out of 100 packets, where each packet contains 2 samples. The results show that our approach preserves diversity, confirming its potential to benefit various related tasks, especially under low-budget conditions.
\begin{table}[]
\centering
\caption{Generalizability of our RD-DPP and Uncertainty-based Decision methods are assessed by selecting samples for the original classification task (\texttt{Smiling}) and using them for two new classification tasks (\texttt{Blond\_Hair} and \texttt{High\_Cheekbones}). Methods are compared in F1 Score and Classification Accuracy.}
\label{tab:multi-task1}
\resizebox{0.7\textwidth}{!}{%
\begin{tabular}{clcccccc} \toprule
\multirow{2}{*}{\textbf{Task}}             & \textbf{Budget} & \multicolumn{2}{c}{\textbf{20}} & \multicolumn{2}{c}{\textbf{40}} & \multicolumn{2}{c}{\textbf{60}} \\ \cline{2-8}
                                           & \textbf{Method}  & \textbf{F1}      & \textbf{ACC}     & \textbf{F1}      & \textbf{ACC}     & \textbf{F1}       & \textbf{ACC}     \\ \midrule
\multirow{3}{*}{\textbf{Smiling}}          & RD-DPP           & \textbf{63.71}   & \textbf{63.97}   & \textbf{70.04}   & \textbf{70.24}   & \textbf{72.45}    & \textbf{72.79}   \\
                                           & Uncertainty Dec.              & 35.64            & 51.51            & 54.29            & 60.38            & 65.75             & 67.58            \\
                                           & $\Delta$         & +28.07            & +12.46            & +15.75            & +9.86             & +6.7               & +5.21             \\ \midrule
\multirow{3}{*}{\textbf{Blond\_Hair}}      & RD-DPP           & \textbf{64.48}   & \textbf{89.5}    & \textbf{66.92}   & \textbf{90.01}   & \textbf{70.41}    & 90.63            \\
                                           & Uncertainty Dec.               & 48.26            & 88.69            & 56.45            & 89.58            & 68.83             & \textbf{90.87}   \\
                                           & $\Delta$         & +16.22            & +0.81             & +10.47            & +0.43             & +1.58              & -0.24            \\ \midrule
\multirow{3}{*}{\textbf{High\_Cheekbones}} & RD-DPP           & \textbf{61.45}   & \textbf{62.68}   & \textbf{67.4}    & \textbf{68.19}   & \textbf{69.17}    & \textbf{70.36}   \\
                                           & Uncertainty Dec.               & 40.47            & 54.32            & 53.88            & 60.82            & 60.84             & 65.11            \\
                                           & $\Delta$         & +20.98            & +8.36             & +13.52            & +7.37             & +8.33              & +5.25            \\ \bottomrule
\end{tabular}%
}
\end{table}

\begin{table}[]
\centering
\caption{The ability (classification accuracy) to resist s negative interference on different tasks.}
\label{tab:multi-task}
\resizebox{0.5\textwidth}{!}{%
\begin{tabular}{ccccc}\toprule
\multicolumn{1}{c}{\textbf{Task}}          & \textbf{Budgets} & \multicolumn{1}{c}{\textbf{10}} & \multicolumn{1}{c}{\textbf{30}} & \multicolumn{1}{c}{\textbf{50}} \\ \midrule
\multirow{3}{*}{\textbf{Rotated MNIST}}    & RD-DPP           & \textbf{53.30}                    &\textbf{ 57.27}                    & \textbf{69.78}                    \\
                                           & Uncertainty Dec.               & 43.09                     & 48.01                    & 66.25                     \\
                                           & $\Delta$         & +10.21                           & +9.26                            & +3.53                            \\ \midrule
\multirow{3}{*}{\textbf{MNIST Fellowship}} & RD-DPP           & \textbf{32.45 }                    & \textbf{55.45 }                    & \textbf{60.88}                    \\
                                           & Uncertainty Dec.               & 17.05                    & 38.29                    & 56.27                    \\
                                           & $\Delta$         & +15.4                            & +17.16                           & +4.61  \\ \bottomrule                         
\end{tabular}%
}
\end{table}

\paragraph{Robustness for Domain-shift Interference:}

Real-world applications often involve data from different sources aiming at similar tasks (domain shift), such as classifying vehicles in urban and rural areas. 
In such cases, training one model for all tasks (i.e. multi-task learning) is not as effective as task-specific models due to the inherent variability among tasks. 
To alleviate this issue, it is advantageous to select samples that preserve task-specific information when switching between different domains under resource constraints~\cite{borsos2020coresets}. We claim that our RD-DPP provides such capability. To this end, we construct two popular synthesis tasks, which are Rotated MNIST \cite{lopez2017gradient} and MNIST Fellowship \cite{douillard2021continuum}\footnote{For more detail about the setup, please refer to Appendix C.3.}. 
The experimental results of the model training on the mixed data and inference on the original task are summarized in Table \ref{tab:multi-task}. The results clearly indicate that the proposed RD-DPP not only outperforms uncertainty-based decisions in addressing the original problem but also demonstrates the capability to reduce the inter-domain interference in multi-task learning.

\section{Conclusion}
Our study reveals a fundamental relationship between the RD and DPP when it comes to selecting diverse training samples to boost the performance of machine learning algorithms. This relationship is used to design a new measure of diversity for data that facilitates sequential DPP inference. We also characterize the \textit{phase transition} property of DPP methods that reveals DPP is more useful for selecting initialization seed points. This is used to design a bi-modal scheduler that switches between the DPP-based and uncertainty-based data selection modes to accommodate different transmission budget constraints better than all alternative selection methods. We showed that our approach can be applied to both raw data and data representation in low-dimensional latent spaces. The intensive experiment results using six different datasets and five different ML/DL models consistently show that our method outperforms pure uncertainty-based, pure diversity-based (including pure DPP-based), and random selection methods. 
Finally, we observed samples selected by our method are more beneficial (compared to other selection methods) for potential future tasks, such as label-shift tasks and domain-shift tasks.

\bibliographystyle{unsrt}  
\bibliography{references}  
\newpage
\appendix
\appendix
\section{\textit{Phase Transition} Property of DPP For Different Distribution}
\begin{figure}[htbp]
\centering\includegraphics[width=0.9\textwidth]{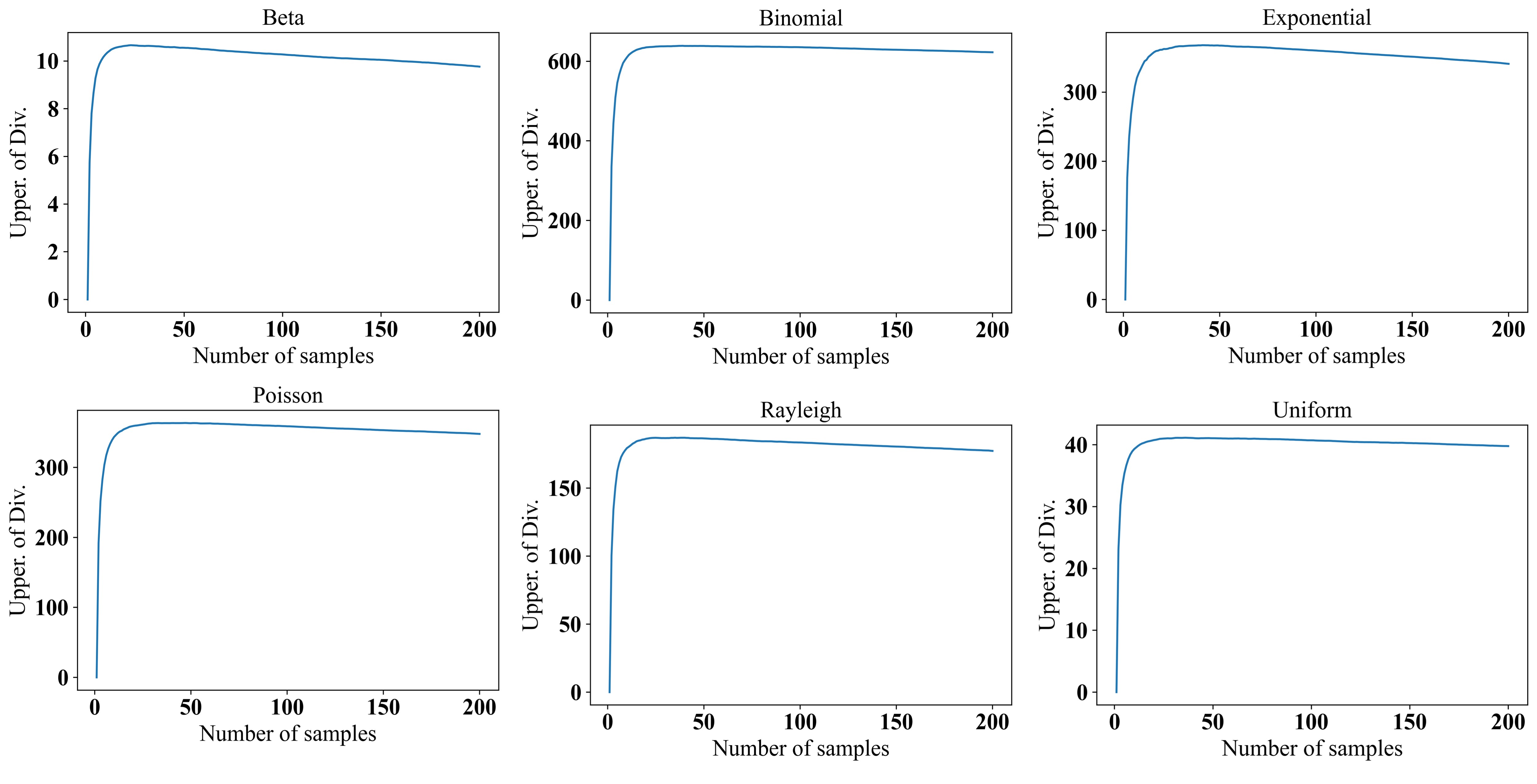}
\caption{The visualization of \textit{phase transition} property of DPP for different distributions.}
\label{fig:view_phase}
\end{figure}

Here we show the \textit{phase transition} property for different distributions in Fig. \ref{fig:view_phase}. 
The distributions include:
\begin{enumerate}
    \item Uniform distribution from $[0,1)$.
    \item Beta distribution with $a=1,b=5$.
    \item Binomial distribution with $n=10, p=0.5$.
    \item Exponential distribution with $\lambda=1$.
    \item Rayleigh distribution with $\sigma=1$.
    \item Poisson distribution with $\lambda=1$.
\end{enumerate}

We sample a dataset from a distribution with 200 samples and 500 dimensions. 

\section{Proof of Eq. (7)}

\begin{proof}
To prove (7), we need to show that $\frac{1}{2} \log \operatorname{det}\left(\mathbf{I}+\alpha \mathbf{Z} \mathbf{Z}^{\top}\right)  = \frac{1}{2} \log \operatorname{det}\left(\mathbf{I}+\alpha \mathbf{Z}^{\top} \mathbf{Z}\right)$ (b), because equality (a) is the definition of RD, and (c) is proven to be the normalization term of DPP.
    Recall the data matrix is $\mathbf{Z}\in\mathbb{R}^{d\times n}$. The Singular Value Decomposition (SVD) of $\mathbf{Z}$ is
    \begin{align}
        \mathbf{Z} = \mathbf{U}_{d\times d}\mathbf{S}_{d\times n} \mathbf{V}^\top_{n\times n},
    \end{align}
    where $\mathbf{U}$ and $\mathbf{V}$ are unitary, and the diagonal elements of $\mathbf{S}$ are the singular values of $\mathbf{Z}$. Suppose the rank of $\mathbf{Z}$ is $r$ meaning that the first $r$ diagonal elements of $S$ are greater than 0. The rest of the diagonal elements and all off-diagonal elements are zero. Hence, 
    \begin{align}
        \mathbf{Z}\mathbf{Z}^\top  =  \mathbf{U}\mathbf{S} \mathbf{V}^\top \mathbf{V}\mathbf{S}^\top \mathbf{U}^T= \mathbf{U}(\mathbf{S} \mathbf{S}^\top) \mathbf{U}^T, \\
        \mathbf{Z}^T\mathbf{Z} =  \mathbf{V}\mathbf{S} \mathbf{U}^\top \mathbf{U}\mathbf{S}^\top \mathbf{V}^T= \mathbf{V}(\mathbf{S}^\top \mathbf{S}) \mathbf{V}^T.
    \end{align}
    Since $\mathbf{U}(\mathbf{S} \mathbf{S}^\top) \mathbf{U}^T$ and $\mathbf{V}(\mathbf{S}^\top \mathbf{S}) \mathbf{V}^T$ are the SVD decomposition of $\mathbf{Z}\mathbf{Z}^\top$ and $\mathbf{Z^\top}\mathbf{Z}$, they have the same non-zero singular values.
    Also, $\mathbf{Z}\mathbf{Z}^\top $ and $\mathbf{Z}\mathbf{Z}^\top $ are \textit{positive semi-definite} (PSD), which implies that their eigenvalues and singular values are the same. Therefore,
    \begin{align}
        \log \operatorname{det}\left(\mathbf{I}+\alpha \mathbf{Z} \mathbf{Z}^{\top}\right) =& \sum_{i=1}^r\log  (\alpha\lambda_i+1)+\sum_{i=1}^{d-r}\log  (1)\\ \nonumber &= \sum_{i=1}^r\log  (\alpha\lambda_i+1), \\
        \log \operatorname{det}\left(\mathbf{I}+\alpha \mathbf{Z}^{\top} \mathbf{Z}\right) &= \sum_{i=1}^r\log  (\alpha\lambda_i+1)+\sum_{i=1}^{n-r}\log  (1) \\ \nonumber &= \sum_{i=1}^r\log  (\alpha\lambda_i+1),
    \end{align}
    where $\lambda_i$ is the $i$th singular value of $\mathbf{Z}\mathbf{Z}^\top$ or $\mathbf{Z^\top}\mathbf{Z}$. Therefore, Eqs. 18 and 19 are equal.
\end{proof}

\section{Packet Preparation and Experiment Setup}

\subsection{Packet Preparation}
In practical systems, the data is transmitted in the form of packets. In our experiment, we assume each packet contains the same number of samples. We also assume data from the same packets are very similar, and we do not operate the intra-packet (i.e. any operations in a packet, such as permutation of the order). 

\begin{figure}[hb]
\centering\includegraphics[width=0.6\columnwidth]{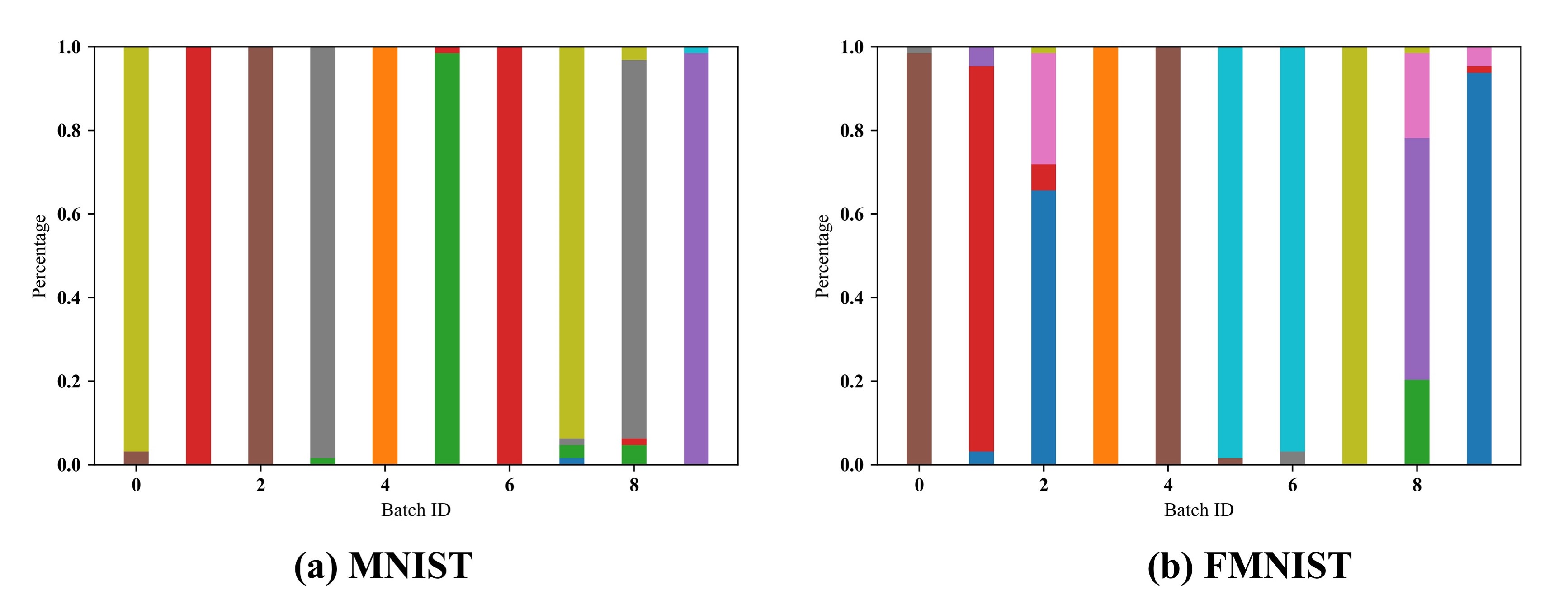}
\caption{The visualization of cluster-based data splitting. Each color denotes a class.}
\label{fig:data_split}
\end{figure}

To generate packets with this assumption, we first use the entire training set to train a random neural network. Then, use 
their representation from this trained network to perform K-means clustering to generate 100 clusters. Noting that generally, the number of samples in each cluster is not the same. Thus, in each experiment, we sample the same number of samples from each cluster (e.g., 64 for MNIST and FMNIST) to encapsulate into a packet, and naturally, we have a total of 100 packets for transmission. We do a similar preparation for CIFAR10 (a total of 100 packets and each packet contains 200 samples) and UCI datasets (a total of 60 packets and each packet contains 5 samples). The composition of some exemplary packets is shown in Fig. \ref{fig:data_split}.

We define the feature of each packet as follows, which is in a relatively fine-grained way and invariant to the order of samples,
\begin{align}\label{eq:feat_fine}
\setlength{\abovedisplayskip}{3pt}
   f(\mathbf{X}_{i}):= \Big[\frac{1}{|C^i_1|}\sum_{e\in {C^i_1}} \mathbf{z}_{e} ,\frac{1}{|C^i_2|}\sum_{e\in {C^i_2}} \mathbf{z}_{e} , \\ \nonumber \cdots,\frac{1}{|C^i_{c_T}|}\sum_{e\in {C_i^{c_T}}} \mathbf{z}_{e}  \Big]\in \mathbb{R}^{dc_T },
   \setlength{\belowdisplayskip}{3pt}
\end{align}
where $C^i_j$ denotes the index set of class $j$ in packet $\mathbf{X}_{i}$, and $\frac{1}{|C^i_j|}\sum_{e\in {C^i_j}} \mathbf{z}_{e}\in \mathbb{R}^{d}$ denotes the averaged features of class $j$. We normalize the obtained vector to have $\| f(\mathbf{X}_{i})\|=1$. 
 
\subsection{Training Detail}

We set $\varepsilon^2=0.5$ in Eqs. (16) for all experiments. 
\subsubsection{MNIST and FMNIST} We set $K=5,\phi_0=2$.
The network architecture for MNIST and FMNIST is presented in Table \ref{tab:NET1}. Each image is mapped to a $288\times 1$ vector)
\begin{table}[H]
\centering
\caption{The architectural details of the network used in MNIST and FMNIST. }
\resizebox{0.4\textwidth}{!}{%
\begin{tabular}{ll}\toprule
Layer Type            & \begin{tabular}[c]{@{}l@{}}\textbf{Kernel Size}\\ $K_1\times K_2\times C_{in}\times C_{out}$\end{tabular} \\ \midrule
Conv2d+ReLU & $3\times3\times 1 \times8$    \\
MaxPool2d             &     -        \\
Conv2d+ReLU & $3\times3\times8\times16$   \\
MaxPool2d             &     -        \\
Conv2d+BatchNorm+ReLU & $3\times3\times16\times32$  \\
MaxPool2d             &     -        \\
Full-connected        & $1\times1\times288\times10$   \\ \bottomrule
\end{tabular}%
}
\label{tab:NET1}
\end{table}
All models use an ADAM optimizer with a learning rate of 1e-3 and a mini-batch size of 64. We train each model with 100 epochs and report the average of the last test accuracy as the final accuracy of the model. The other experiments use the same way to report. 

\subsubsection{CIFAR10}
We set $K=10$. We use SGD optimizer in this experiment. We set the learning rate to 0.01, the momentum factor to 0.9, and the weight decay factor to 5e-4. We also use a cosine annealing schedule and set $T_{max}$ to 200.


\subsubsection{UCI Datasets}
All datasets are split into 70\%-30\% training and test subsets and pre-processed by Z-score normalization. We set $K=3$. The learning rate was set to 1e-2, and 1e-3 for Yeast, Cardiotocography, and Statlog, respectively.

\subsubsection{Linear Evaluation Protocol}
All models use an ADAM optimizer with a learning rate of 1e-3 and a mini-batch size of 64.

\subsection{Datasets Setup of Multi-task Learning}
\paragraph{Rotated MNIST:} In each sub-task, the digits were rotated by a pre-defined angle. Each task in Rotation MNIST is a 10-class classification problem where their labels are the corresponding digits. Thus, each subsequent task involves classification on the same ten digits.
\paragraph{MNIST Fellowship:} The MNIST Fellowship is a combination of MNIST, Fashion MNIST, and  KMNIST. Each sub-task corresponds to one dataset with ten classes of annotation. This task has more various domain differences than the previous one.

Each of these two tasks consists of three sub-tasks with MNIST digits classification being the first sub-task. Specifically, we apply different selection strategies to the first sub-task (original task), maintaining consistency with our previous experiment (as shown in Table 1). For this experiment, we set a budget of 10-50, and each budget has 64 samples. Subsequently, an equal number of samples are randomly sampled from the remaining two sub-tasks.

\subsection{Hardware}
All experiments were implemented based on Pytorch Framework and on a cluster with an Intel(R) Xeon(R) Gold 6148 CPU with 125GB of memory and an NVIDIA A100 with 80GB of memory.

\end{document}